\documentclass[10pt,twocolumn,letterpaper]{article}

\usepackage{cvpr}      
\usepackage{graphicx}
\usepackage{amsmath}
\usepackage{amssymb}
\usepackage{booktabs}

\usepackage[pagebackref,breaklinks,colorlinks]{hyperref}

\usepackage{multirow}
\usepackage{graphicx}
\usepackage{bbding}
\usepackage{array}
\makeatletter

\usepackage[capitalize]{cleveref}
\crefname{section}{Sec.}{Secs.}
\Crefname{section}{Section}{Sections}
\Crefname{table}{Table}{Tables}
\crefname{table}{Tab.}{Tabs.}

\begin{document}

\title{Mutual Balancing in State-Object Components for Compositional Zero-Shot Learning }

\author{
	Chenyi Jiang\textsuperscript{\rm 1}, Dubing Chen\textsuperscript{\rm 1}, Shidong Wang\textsuperscript{\rm 2}, Yuming Shen\textsuperscript{\rm 3}, Haofeng Zhang\textsuperscript{$1$ *}, Ling Shao\textsuperscript{\rm 4 }\\
	\textsuperscript{\rm 1}Nanjing University of Science and Technology, 
	\textsuperscript{\rm 2}University of Newcastle-upon-Tyne, \\
	\textsuperscript{\rm 3}University of Oxford, 
	\textsuperscript{\rm 4}Terminus Group, Beijing, China\\
}
\maketitle

%%%%%%%%% ABSTRACT
\begin{abstract}
Compositional Zero-Shot Learning (CZSL) aims to recognize unseen compositions from seen states and objects. The disparity between the manually labeled semantic information and its actual visual features causes a significant imbalance of visual deviation in the distribution of various object classes and state classes, which is ignored by existing methods. To ameliorate these issues, we consider the CZSL task as an unbalanced multi-label classification task and propose a novel method called \textbf{MU}tual balancing in \textbf{ST}ate-object components (\textbf{MUST}) for CZSL, which provides a balancing inductive bias for the model. In particular, we split the classification of the composition classes into two consecutive processes to analyze the entanglement of the two components to get additional knowledge in advance, which reflects the degree of visual deviation between the two components. We use the knowledge gained to modify the model's training process in order to generate more distinct class borders for classes with significant visual deviations. Extensive experiments demonstrate that our approach significantly outperforms the state-of-the-art on MIT-States, UT-Zappos, and C-GQA when combined with the basic CZSL frameworks, and it can improve various CZSL frameworks. Our codes are available on \url{https://anonymous.4open.science/r/MUST_CGE/}
\end{abstract}

\section{Introduction}
\label{sec:intro}
Humans are extremely superior to computers at interpreting new information and expanding contemporary knowledge. A fascinating example is that an infant can perceive the concept of a ``red apple'' once it has experienced recognizing a ``red tomato'' and a ``green apple''. Giving computers this knowledge-transfer capability is vitally important because creating complete composited data is a costly endeavor \cite{salakhutdinov2011learning,wang2017learning}. Compositional Zero-Shot Learning (CZSL), in this case, has been proposed to teach computers this knowledge transfer ability to distinguish unseen compositions of seen components, \eg, objects and states. 

The methods based on CZSL can be roughly summarized into two categories. The first group of methods usually trains specialized state and object classifiers to transform the CZSL task into a standard supervised recognition task \cite{li2020symmetry,misra2017red}. However, learning a transformation network based on an individual classifier is difficult to capture the variations in visual representations caused by different compositions which in turn makes it extremely challenging to create high-accuracy models. 

\begin{figure}[t]
	\centering
	\includegraphics[width=1.00\columnwidth]{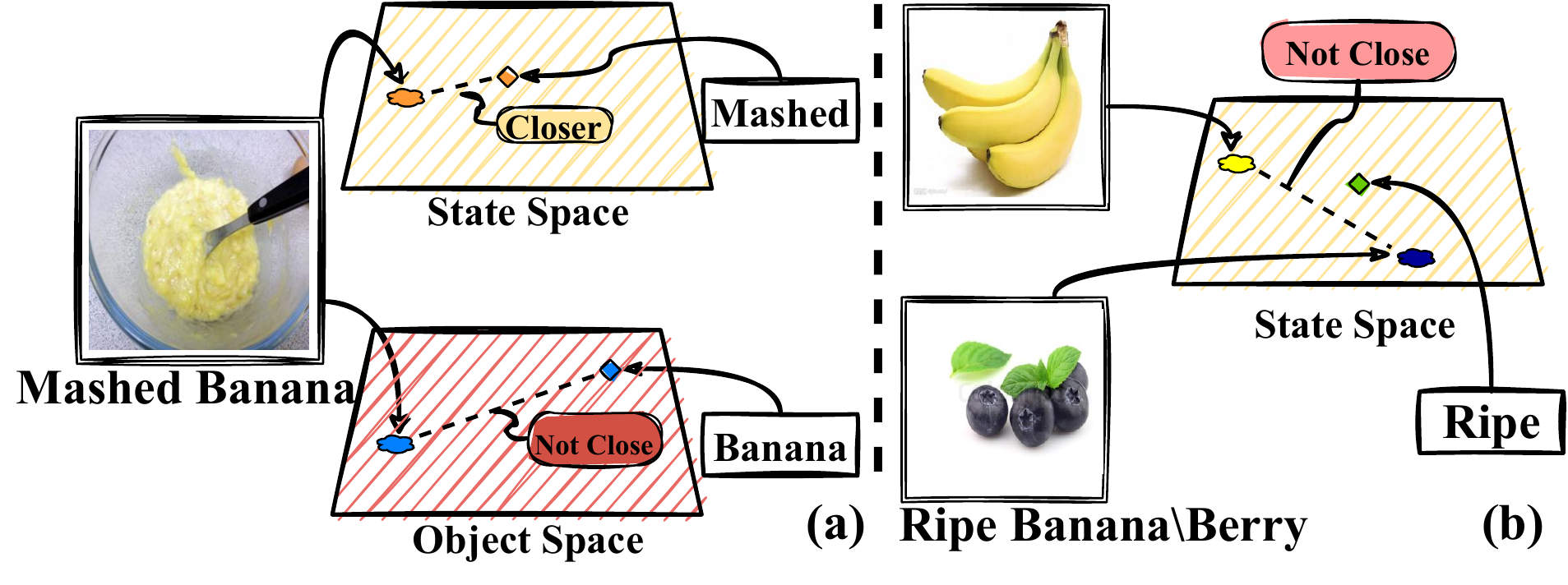} 
	\caption{Motivation for our method. \textbf{(a)} ``Mashed'' significantly changes the appearance of the visual features of the ``banana''. \textbf{(b)} Compositionality allows the visual features of ``ripe'' to appear in various compositions in different colors.}
	\label{fig1}
\end{figure}
The second group of methods can effectively treat each composition as a unique class by projecting state-object compositions into a common joint embedding space. For example, a graph convolutional neural network (GCN) is regarded as a joint embedding function to establish dependencies between the state and the object \cite{naeem2021learning}. Despite the success of existing methods, they are prone to collapse due to overlooking the following two unresolved problems:

\begin{figure*}[ht]
	\centering
	\includegraphics[width=0.80\textwidth]{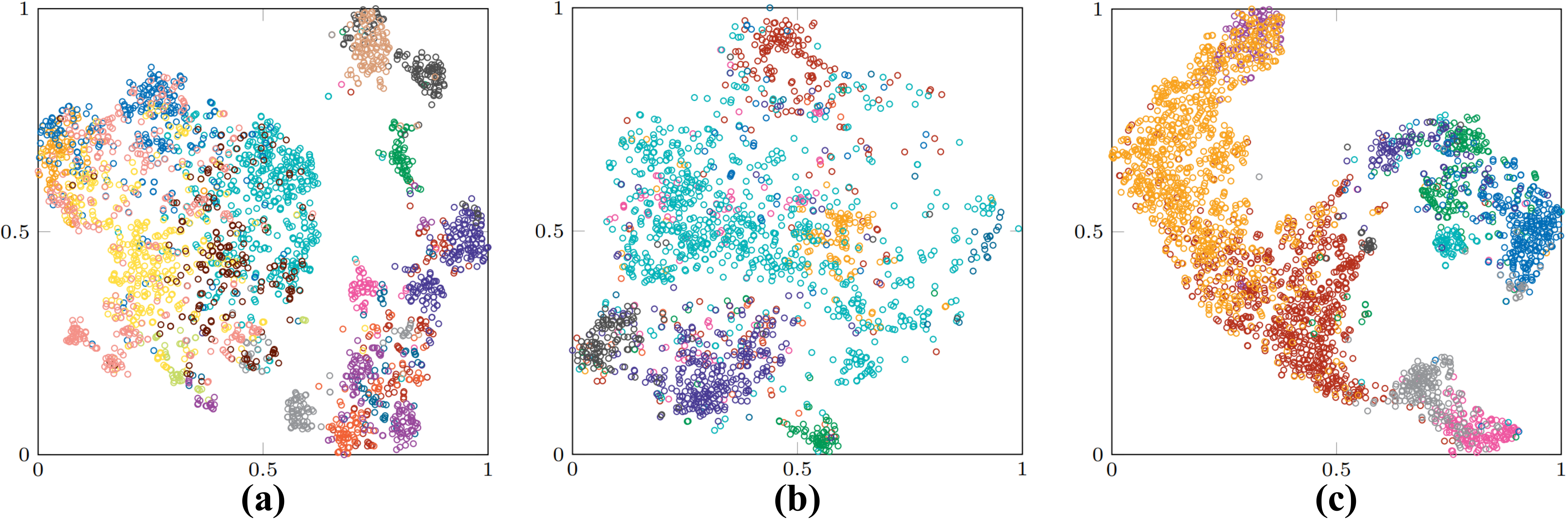} 
    \vspace{-1ex}
	\caption{Visualization of the visual features on UT-Zappos \cite{utzappos}. (\textbf{a}) Visual features are extracted using a pre-trained ResNet18 backbone and visualized with the t-sne \cite{tsne}. Distinct composition classes are presented in different colors. (\textbf{b}) and (\textbf{c}) are the features in terms of state and object separated from visual features using MLP layers and visualized with t-sne, where the different colors are used to indicate different states or object classes. (\textbf{a}) shows that visual features lack classification boundaries, whereas (\textbf{b}) and (\textbf{c}) illustrate the varying degrees of visual deviation that are generated by compositional effects on objects and states, resulting in the imbalances between them.}
	\label{fig6}
\end{figure*}

\textit{(1) Inconsistent visual deviations of the two components within the composition.} Compositionality may cause the deviation issue in the sample's visual representation, but in reality, the deviation is inconsistent across all components. As shown in Fig. \ref{fig1} (\textbf{a}), the features of ``mashed'' in the ``mashed banana'' sample are closer to word vectors encoded according to their labels, while the compositionally simply introduces the concepts of the ``yellow'', ``mushy'' and \etc. However, the visual deviation of the \textit{banana} features is tremendous as they completely lose the shape and texture of the ``banana''.

\textit{(2) The visual deviation of the same component varies between compositions.} Intuitively, the same component in different compositions should have identical or at least similar expressions, but the reality is totally different. In the phrases ``ripe banana'' and ``ripe berry'', as shown in Fig. \ref{fig1} (\textbf{b}), both contain the ``ripe'' component, but it appears in distinct colors, \ie, the ``banana'' is shown in yellow and ``berry'' is displayed in dark blue. If CZSL is viewed as a multi-label task with specific constraints, \ie, objects and states as separate classes, such features significantly increase the intra-class distance of the ``ripe'' class thus harming the performance of CZSL.

In fact, if CZSL is considered as a multi-label task with certain constraints, then the two imbalance problems mentioned above will produce the same result, \ie, \textit{the visual deviation caused by compositionality to the components is not constant, which is referred as the component imbalance in CZSL}. Is it possible to make a classifier explicitly aware of component imbalance information during training without increasing the complexity of the backbone network for feature extraction? In this paper, we consider alleviating the impact of these two issues by proposing an object-state balancing method. Concretely, it can be achieved by (1) introducing the concept of the additional information derived from the degree of visual deviation between the two components, and (2) then making use of the additional knowledge during training through re-weighting the vanilla cross-entropy loss. This approach, named a Mutual Balancing in State-Object Components network (MUST), realizes more balanced results with a fixed backbone. 

The key contributions of this approach are summarized as follows:
\begin{itemize}
  	
\item Innovatively considering the CZSL task as an unbalanced multi-label classification task, our method utilizing the visual deviation of the two components to evaluate their imbalance, to provide an inductive bias for the model.
\item We used the imbalance information among the components to re-weight the training process of CZSL, allowing the model could reconstructs the inter-components balance relationship on this basis. 
\item Our proposed strategy significantly reduces the imbalance between components. We test our approach on three challenging datasets; it outperforms SoTAs when combined with the base CZSL methods and can be a plug-in to augment various joint embedding function-based methods.
\end{itemize}

\section{Related Work}
\textbf{Compositional Zero-Shot Learning. } The knowledge transfer ability of models has been the core exploration of Zero-Shot Learning (ZSL), \ie learning generalizable visual-semantic relations from the seen classes and projecting the encoded relations into a common space to facilitate effective linking with the unseen classes \cite{lampert2013attribute,lampert2009learning,akata2013label,elhoseiny2013write}. In contrast, Compositional Zero-Shot Learning (CZSL) builds on this foundation and focuses more on examining aspects of the sample's compositions \cite{misra2017red}. More specifically, given a training set in which each sample consists of a composition of objects and states, CZSL requires to identify new compositions derived from the seen objects and the states in the testing stage. 

Existing methods can be broadly classified into two groups. The first group of methods \cite{yang2020learning , chen2014inferring} follow a unique pipeline by introducing two classifiers to independently identify the prototypes of the object and state classes, and then assemble their results \cite{hoffman1984parts,biederman1987recognition}. Concretely, \cite{chen2014inferring} proposes a tensor factorization approach that infers unseen object-state pairs using a sparse set of class-specific SVM classifiers trained with seen compositions. \cite{nagarajan2018attributes} uses learned semantic embeddings to remove states from sample objects and a regularizer to convey state effects better. Although these methods have sought to learn the compositionality from samples, the assembled objects and states contain significant visual deviations compared to a typical supervised task, thus hindering the effectiveness to achieve satisfactory performance. 

The second group of approaches aim to realize the classification by learning a joint embedding function from the corresponding images, objects and states \cite{purushwalkam2019task, wang2019task,atzmon2020causal}. For example, CGE \cite{naeem2021learning} establishes the dependencies between objects and states using a Graph Convolutional Neural (GCN) network. Furthermore, \cite{li2020symmetry} exploits the symmetrical relationships between states and objects to create better prototypes. A contrastive-based learning approach is proposed by \cite{li2022siamese} to produce better generalization ability for new compositions by isolating more primitive class prototypes.

However, if the objects and states are regarded as independent classes, then their degree of visual deviation in the dataset is unbalanced, meaning that a number of objects and states might have relatively weak visual representations, while the rest may have substantial intra-class compositional differences.

\textbf{Imbalanced Training Data.} Many studies have been devoted to solving the problem of sample imbalance between classes \cite{lin2017focal,liu2016ssd,redmon2017yolo9000,zhou2017fine} as the unbalanced training sample can significantly affect the performance of the model \cite{masko2015impact}. \cite{redmon2017yolo9000} employs the structural relationships of the labels to classify the labels into levels, performing data augmentation on each level separately and thus retaining a balance between different classes. To alleviate the impact of the Easy Sample Domain problem, \cite{zhou2017fine} proposes an approach of active incremental learning, which encourages the model to actively learn more difficult classes first. \cite{lin2017focal} integrates the difficulty of each sample with an objective function to adaptively evaluate the difficulty of the sample at each iteration to strengthen the learning for those hard samples. 

However, none of the above methods can be directly transferred to the CZSL problem because the imbalance in CZSL does not originate entirely from the classes but partly from the compositional components of the samples. Our proposed method lies at the intersection of several previously discussed approaches. Concretely, we treat the CZSL problem as an unbalanced multi-label task with specific constraints and promote the model to focus more on samples containing objects or states with unpredictable distributions. To this end, an additional module is proposed to explore the visual deviations of components in advance, and then the components with smaller visual deviations are introduced into the final result as additional knowledge while the components with larger visual deviations are used as the moderators of the objective function. 

\begin{figure}[t]
	\centering
	\includegraphics[width=1.00\columnwidth]{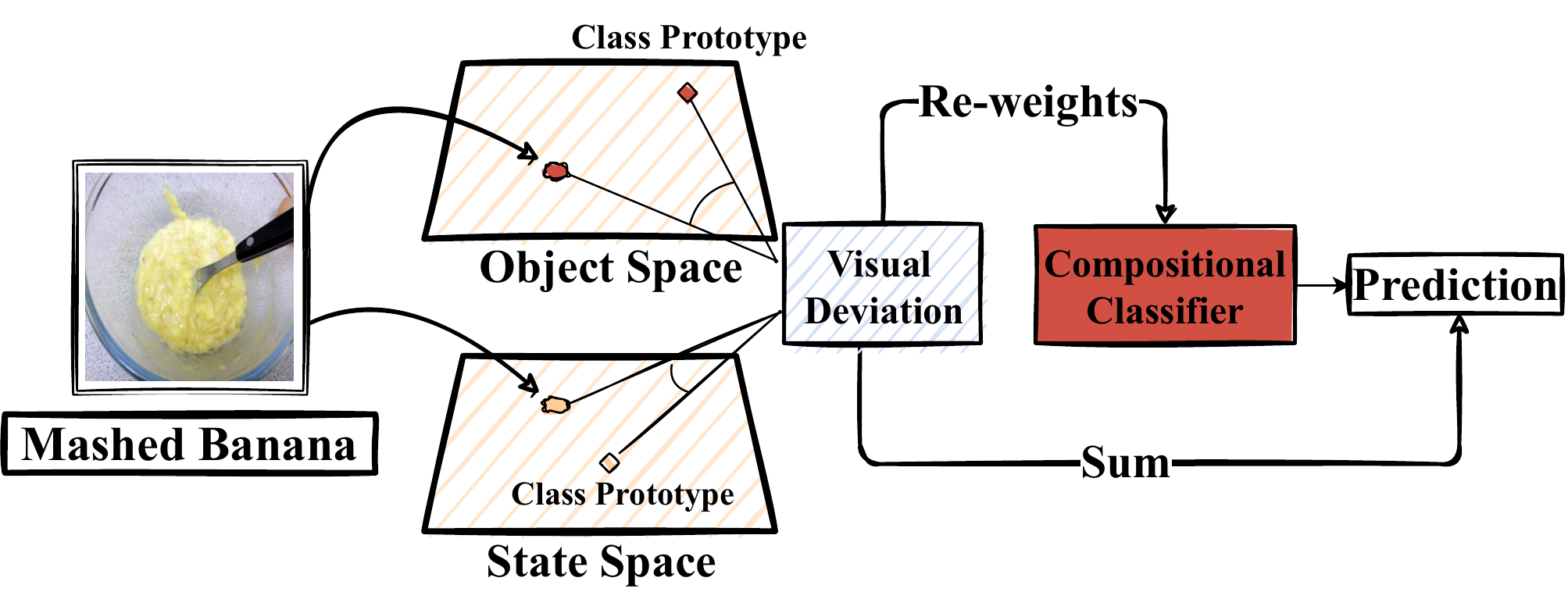} 
	\caption{Illustration of the core idea. Visual features are deconstructed and projected into the corresponding space, \ie, the object space and state space, and then their visual deviations are computed accordingly based on their similarities to the class prototypes. We use the visual deviations to roughly quantify the balance between the components, insert them as adjustable weights into subsequent classifiers, and output them as extra adjustments to the final result. }
	\label{fig2}
\end{figure}

\section{Approach}
CZSL needs reliable generalization-type knowledge extracted from seen objects and states as well as their compositions, such as "white cat" and "black dog," and extends the resulting knowledge to the new compositions of unseen classes, such as "black cat." This is a challenging task because the visual deviations of the different components in the data are not balanced and there is no certain pattern of variations for the other compositions. 

\subsection{Empirical Analysis on Visual Features}
\label{sec.3.2}

\begin{figure*}[ht]
	\centering
	\includegraphics[width=0.85\textwidth]{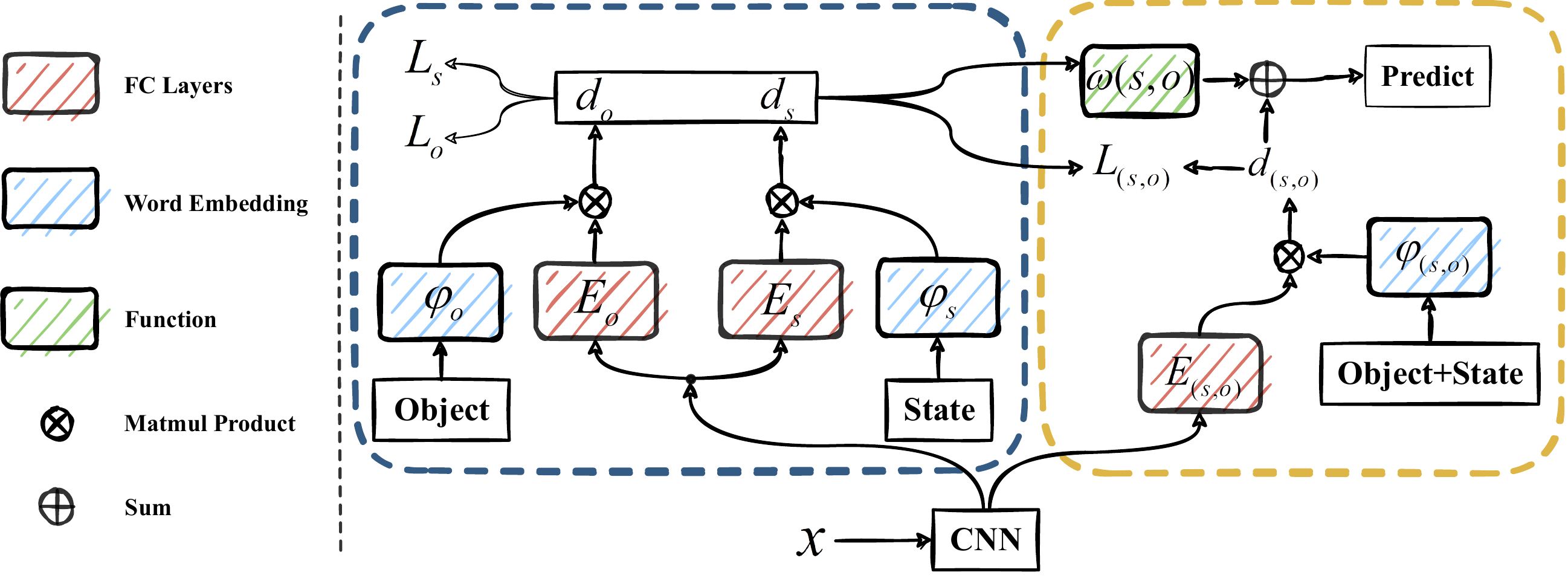} 
	\caption{Illustration of the proposed framework. $E_{o}$ and $E_{s}$ are two embedding functions used for extracting features of objects and states. Once the visual features pass $E_{o}$ and $E_{s}$, calculate the cosine similarities with the word vectors \cite{mikolov2013distributed} to generate $d_{s}$ and $d_{o}$, where $d_{s}$ and $d_{o}$ are used to estimate the amount of visual deviations involved in training through re-weighting the vanilla cross-entropy loss. Noteworthy, $d_{s}$, $d_{o}$ also provide additional information for the final classification results according to Eq. \ref{eq_9}; During inference, the visual features are processed via the embedding function $E_{(s,o)}$, and only the cosine similarity of $d_{(s,o)}$ needs to be calculated with the word vector. The objective function $\mathcal{L}_{s,o}$ can selectively adjust the weights of each sample according to $d_{s}$ and $d_{o}$, which will be discussed in Sec. \ref{sec.3.3}. }
	\label{fig3}
\end{figure*}

In CZSL, one of the most challenging tasks is to establish reliable visual-semantic relations. The t-sne \cite{tsne} visualization of visual features extracted from the test set of UT-Zappos \cite{utzappos}, shown in Fig. \ref{fig6} (\textbf{a}), reflects that the absence of clear and linear boundaries between most classes. Two primary reasons for such problems are: \textit{ 1) some objects and states over-dominate the visual features of the samples, blurring the boundaries between classes}, and \textit{2) various deviations of the manual semantic information from sample visual representations.}  Although the lack of fine-grained discriminative ability of the backbone also contributes to this phenomenon, it is orthogonal to our study and will not be discussed in depth. 

We also use MLP layers to retrieve the object and state features from visual features and then visualize the generated results. In other words, state and object are treated as independent classes. As shown in Fig. \ref{fig6} (\textbf{b}) and (\textbf{c}), it reveals the following facts: \textit{1) clear boundaries have arisen for some categories although there are a few classes where the boundaries are not linearly separable}, and 2) \textit{different objects and states produce diverse visual deviations affected by the compositionality.} This motivates us to incorporate the unbalanced relationships between objects and states into the training by re-weighting the vanilla cross-entropy loss.

To cope with the above issues, it introduces new modules to capture the interrelations between unbalanced components of datasets that are inherently unavailable. Concretely, it deconstructs and maps visual features into a new space, and then estimates the similarity to the corresponding class prototype which induces components with large visual deviations and components with small visual deviations. The quantities of the resulting two visual deviations can then be used to approximate the balance between components. An example is illustrated in In Fig. \ref{fig2}, where the ``mashed" and ``banana" are used to depict the given sample. Assuming that the ``mashed" corresponds to the component with less visual deviation and ``banana" is the more significant one. The previously received mutual balance relationship is introduced into the training of the model to force the model to distance itself from other similar classes.

\subsection{Problem Definition}
In CZSL, each sample is composed of two primary components, including the states and objects. We use $\mathcal{S}$ and $\mathcal{O}$ to denote the set of objects and the set of states, respectively. $\mathcal{C}$ is used to represent the set of compositions, \ie, $\mathcal{C}=\mathcal{S}\times\mathcal{O}=\left \{(s,o)|s\in \mathcal{S}, o\in \mathcal{O}\right \} $. Only those natural compositions of $\mathcal{C}$ are used in the closed world, \ie, $\mathcal{C}_{closed} \subseteq  \mathcal{C} $. The seen set is denoted as $\mathcal{T}=\left \{(x,y)|x\in \mathcal{X}_{t}, y\in \mathcal{Y}_{t} \right \}$, where $\mathcal{X}_{t}$ represents seen images, $\mathcal{Y}_{t}\subset \mathcal{C}_{closed}$ is the label of training images, and $y = (s, o)$. On the other hand, the unseen set is denoted as $\mathcal{N}=\left \{(x,y)|x\in \mathcal{X}_{n}, y\in \mathcal{Y}_{n} \right \}$, implying that $\mathcal{X}_{t} \cap \mathcal{X}_{n} =$\O, $\mathcal{Y}_{t}\cup \mathcal{Y}_{n} = \mathcal{C}_{closed}$ and $\mathcal{Y}_{t} \cap \mathcal{Y}_{n}  = $\O. We approach the CZSL problem according to the setting of Generalized Zero-Shot Learning, \ie, the test set contains seen and unseen compositions, which requires learning a mapping function $\mathcal{X}\rightarrow \mathcal{C}_{closed}$.

\subsection{Mutual Balance in State-Object Components}\label{sec.3.3}
As mentioned in Sec. \ref{sec:intro}, CZSL can be viewed as an imbalanced multi-label task under certain constraints, where the imbalances are reflected in varying degrees of visual deviations between components. The theoretical distance between visual features and class prototypes is proposed to approximate the visual deviation existing in two components, but overly training the model might yield overfitting and losing such deviant information. Therefore, the model is needed to preserve the deviant information to the greatest extent during training. 

The above idea can be realized by means of implementing additional classifiers for the components. As shown in Fig. \ref{fig3}, two embedding functions $E_{o}$ and $E_{s}$ (boxes colored in blue) are used to separate objects and states in visual features, denoted as $\mathbf{h}_{o},\mathbf{h}_{s}\in \mathbb{R}^{k}$, where $k$ is the dimension of the joint embedding space. This space also contains the word vectors that encode the labels of components, which are denoted as $\varphi_{s}(s) \in \mathbb{R}^{k}$ and $\varphi_{o}(o) \in \mathbb{R}^{k}$. Then, the classification results are yielded by calculating the cosine similarity, $d_{o} = cos(\mathbf{h}_{o}, \varphi_{o}(o))$ , $d_{s}=cos(\mathbf{h}_{s}, \varphi_{s}(s))$, which are also used to estimate the degrees of visual deviations.
Specifically, the focal point of the model can be improved according to the dependencies between objects and states. Formally, the dependencies can be denoted as: 
\begin{equation}\label{eq_3}
\psi(s,o)=\left\{
\begin{array}{rcl}
1 & & {(s,o) \in \mathcal{Y}_{t},}\\
0 & & {else.}\\
\end{array} \right.
\end{equation}

This facilitates defining $\bar{d}_{s}^{o}=\psi(s,o)d_{s}$ and $\bar{d}_{o}^{s}=\psi(s,o)d_{o}$ for constructing the conversion relationships between an individual component and other components. Based on the resulting dependencies, a novel objective function is proposed to achieve two goals: 1) \textit{Reducing the weight of samples with less visual deviations in the two components during optimization}, and 2) \textit{Implementing the model optimization process with higher bias, \ie, to maintain the discrepancies in the degree of visual deviations between the two components.} Inspired by \cite{lin2017focal}, we amend the standard cross-entropy loss and introduce one of our objective functions:
\begin{equation}\label{eq_4}
    \begin{aligned}
   \mathcal{L}_{o}= -(1-\bar{d}_{s}^{o})^{\gamma}log\frac{e^{cos(\mathbf{h}_{o}, \varphi_{o}(o))}}{\sum_{\hat{o}\in \mathcal{O}} e^{cos(\mathbf{h}_{o}, \varphi_{o}(\hat{o}))} },
    \end{aligned}
\end{equation}
\begin{equation}\label{eq_5}
    \begin{aligned}
   \mathcal{L}_{s}= -(1-\bar{d}_{o}^{s})^{\gamma}log\frac{e^{cos(\mathbf{h}_{s}, \varphi_{s}(s))}}{\sum_{\hat{s}\in \mathcal{S}} e^{cos(\mathbf{h}_{s}, \varphi_{s}(\hat{s}))} },
    \end{aligned}
\end{equation}
where $\gamma$ is a hyperparameter that determines the effect of $\bar{d}_{s}^{o}$ and $\bar{d}_{o}^{s}$. 

The implementation of the above-introduced module realizes the determining of the degree of balance between two sample components, but the CZSL is still limited by other issues. Then, it aims to explore how to make the most of the received imbalance information to re-weight the subsequent training, \ie, to shift the focus of the model, as displayed by the yellow box in Fig. \ref{fig3}. Concretely, we construct an embedding function $E_{(s, o)}$ for compositional classification and represent the compositional visual features by $\mathbf{h}_{(s,o)}\in \mathbb{R}^{k}$. The compositional labels, denoted by $(s,o)$, are also projected into a joint embedding space of $\varphi_{(s, o}(s,o) \in \mathbb{R}^{k}$. Then, the cosine similarities between the visual features $\mathbf{h}_{(s, o)}$ and word vectors $\varphi_{(s, o}(s,o)$ are calculated by $d_{(s,o)}=cos(\mathbf{h}_{(s, o)}, \varphi_{(s,o)}(s, o))$.

In order to reach the best balance of the predictions for these two compositional components, we contemporarily introduce $d_{s}, d_{o}$ for model training. To this end, it needs to build the dependencies between the components and compositions by using: 
\begin{equation}\label{eq_6}
\hat{\psi}(a,y)=\left\{
\begin{array}{rcl}
1 & & {a \in y.}\\
0 & & {else.}\\
\end{array} \right. ,
\end{equation}
where $a$ is flexible to be $s$ or $o$, and $y=(s,o)$ is the compositional label. The conversion relationships between a single component to compositions are constructed and denoted as $\hat{d}_{a}^{y}=\hat{\psi}(a, y)d_{a}$. The compositional classifier will place more emphasis on samples that have performed relatively unbalanced in previous predictions. Formally, it can be expressed as:
\begin{equation}\label{eq_7}
\begin{split}
    \begin{aligned}
   \mathcal{L}_{(s,o)}&= -\mu(s,o)^{\gamma} log\frac{e^{cos(\mathbf{h}_{(s,o)}, \varphi_{(s,o)}(s,o))}}{\sum_{(\hat{s},\hat{o}) \in \mathcal{Y}_{t}} e^{cos(\mathbf{h}_{(s,o)}, \varphi_{(s,o)}(\hat{s},\hat{o}))} },\\
    \end{aligned}
\end{split}
\end{equation}
where $\mu(s,o)=(1-\hat{d}_{s}^{(s,o)})(1-\hat{d}_{o}^{(s,o)}),$ is the re-weighting factor derived from $d_{s}$ and $d_{o}$. During training, samples are re-weighted according to the degree of balance between the components to induce the model to generate sharper classification boundaries for those categories with significant visual deviations. 
\subsection{Training and Inference}
Overall, our model is optimized with the following objective function:
\begin{equation}\label{eq_8}
    \begin{aligned}
   \mathcal{L}= \mathcal{L}_{(s,o)}+\lambda(\mathcal{L}_{s}+\mathcal{L}_{o}),
    \end{aligned}
\end{equation}
where $\lambda$ is a hyper-parameter introduced to reduce perturbations caused by two different objective functions. During inference, we first compute the cosine similarities between the class prototypes of the objects and states, denoted by $\mathbf{h}_{o}$ and $\mathbf{h}_{s}$ in order to acquire $d_{s}$ and $d_{o}$. Then, the resulting $d_{s}$ and $d_{o}$ are incorporated into the final classification results to enhance the knowledge regarding the visual deviations. This can make the classification results more favorable for specific components with less deviations, further restraining the interference caused by the compositionality. This process can be expressed as:
\begin{equation}\label{eq_9}
    \begin{aligned}
   \omega(s,o)= \frac{\underset{\hat{o}\in \mathcal{O},\hat{s}\in \mathcal{S}}{max}\hat{d}_{s}^{(\hat{s},\hat{o})}}{\underset{\hat{o}\in \mathcal{O},\hat{s}\in \mathcal{S}}{max}\hat{d}_{s}^{(\hat{s},\hat{o})}+\underset{\hat{s}\in \mathcal{S},\hat{o}\in \mathcal{O}}{max}\hat{d}_{o}^{(\hat{s},\hat{o})}}.
    \end{aligned}
\end{equation}

we approximate the confidence of the model predictions by exploiting the maximum value for each class. The goal of this function is to logically combine the two predictions based on their confidence and append them to the final classification result. In this way, it is beneficial to make more confident predictions about the components when the final classification is conducted. Finally, the inference rule is described as:
\begin{equation}\label{eq_10}
 \resizebox{1.0\linewidth}{!}{$
    \begin{aligned}
  \tilde{y}=\underset{(s,o)\in \mathcal{C}_{closed}}{argmax}[\omega(s,o)\hat{d}_{s}^{(s,o)}+(1-\omega(s,o))\hat{d}_{o}^{(s,o)}+d_{(s,o)}].
    \end{aligned}
   $}
\end{equation}

\begin{table*}[t] 
		\centering
		\resizebox{\textwidth}{!}{
			\begin{tabular}{@{}l|cccccc|cccccc|cccccc}
				\toprule
		\multirow{3}{*}{Method} &\multicolumn{6}{c|}{MIT-States} & \multicolumn{6}{c|}{UT-Zappos}& \multicolumn{6}{c}{C-GQA} \\ 
        &\multicolumn{1}{c}{AUC}&\multicolumn{3}{c}{Best}& & &\multicolumn{1}{c}{AUC}&\multicolumn{3}{c}{Best}& & &\multicolumn{1}{c}{AUC}&\multicolumn{3}{c}{Best}& & \\
        &\multicolumn{1}{c}{Test}&$\mathcal{A}^{H}$&$\mathcal{A}^{S}$&$\mathcal{A}^{U}$&$\mathcal{A}^{adj}$&$\mathcal{A}^{obj}$&\multicolumn{1}{c}{Test}&$\mathcal{A}^{H}$&$\mathcal{A}^{S}$&$\mathcal{A}^{U}$&$\mathcal{A}^{adj}$&$\mathcal{A}^{obj}$&\multicolumn{1}{c}{Test}&$\mathcal{A}^{H}$&$\mathcal{A}^{S}$&$\mathcal{A}^{U}$&$\mathcal{A}^{adj}$&$\mathcal{A}^{obj}$\\
				\midrule
        AttOp \cite{nagarajan2018attributes}&1.6&9.9&14.3&17.4&21.1&23.6&25.9&40.8&59.8&54.2&38.9&69.6&0.7&5.9&17.0&5.6&-&-\\
LE+ \cite{misra2017red}&2.0&10.7&15.0&20.1&23.5&26.3&25.7&41.0&53.0&61.9&41.2&69.2&0.8&6.1&18.1&5.6&-&-\\
TMN \cite{purushwalkam2019task}&2.9&13.0&20.2&20.1&23.3&26.5&29.3&45.0&58.7&60.0&40.8&69.9&1.1&7.5&23.1&6.5&-&-\\
SymNet \cite{li2020symmetry} &3.0&16.1&24.4&25.2&26.3&28.3&23.9&39.2&53.3&57.9&40.5&71.2&2.1&11.0&26.8&10.3&-&-\\
SCEN* \cite{li2022siamese}&5.3&18.4&29.9&25.2&28.2&32.2&32.0&47.8&63.5&63.1&\textbf{47.3}&\textbf{75.6}&2.9&12.4&28.9&12.1&\textbf{13.6}&27.9\\
\midrule
CompCos \cite{mancini2021open} &4.5&16.4&25.3&24.6&27.9&31.8&28.7&43.1&59.8&62.5&44.7&73.5&2.6&12.4&28.1&11.2&-&-\\
CompCos+\textbf{MUST} (Ours)&5.6&18.7&27.9&27.2&\textbf{30.3}&\textbf{33.9}&31.7&47.5&\textbf{63.6}&63.5&46.2&73.6&\textbf{3.0}&\textbf{13.6}&27.8&\textbf{13.2}&10.4&29.4\\
\midrule
CGE \cite{naeem2021learning} &5.1&17.2&28.7&25.3&27.9&32.0&26.4&41.2&56.8&63.6&45.0&73.9&2.3&11.4&28.1&10.1&-&-\\
CGE+\textbf{MUST} (Ours)  &\textbf{6.0}&\textbf{19.5}&\textbf{30.1}&\textbf{26.3}&28.7&33.6&\textbf{32.8}&\textbf{48.4}&60.7&\textbf{64.7 }&46.3&73.3&2.9&13.1&\textbf{29.8}&12.5&11.5&\textbf{30.6}\\
	\bottomrule
			\end{tabular}
}
		\caption{The performance of the CZSL compared with the state-of-the-art methods, where $\mathcal{A}^{adj}$, $\mathcal{A}^{obj}$, $\mathcal{A}^U$, $\mathcal{A}^S$ and $\mathcal{A}^H$ represent the Top-1 accuracy in \ref{sec.4.1}. $AUC (in \%)$ is summarized by performing the model on the test set. The test data on C-GQA are from \cite{mancini2022learning}. The (*) indicates that C-GQA utilized by SCEN \cite{li2022siamese} is slightly different than the other methods. The best results are shown in bold, \textbf{MUST} is the proposed method.}
	\label{tab.1}
    \end{table*}

\begin{table}[ht]
\scalebox{0.65}{
	\centering
    \begin{tabular}{@{}lcccccccccc}
    \toprule
		\multirow{2}{*}{Dataset} & & & \multicolumn{2}{|c|}{Training}   & \multicolumn{3}{c|}{Validation}   & \multicolumn{3}{c}{Test} \\
      & $s$ & $o$   &  \multicolumn{1}{|c}{$sp$}  & $i$& \multicolumn{1}{|c}{$sp$} & \multicolumn{1}{c}{$up$} &\multicolumn{1}{c|}{$i$}  & \multicolumn{1}{c}{$sp$} & \multicolumn{1}{c}{$up$} & \multicolumn{1}{c}{$i$}\\
				\midrule
		MIT-States \cite{mit} &115&245&1262&30k&300&300&10k&400&400&13k \\
	    UT-Zappos \cite{utzappos} &16&12&83&23k&15&15&3k&18&18&3k \\
	    C-GQA\cite{naeem2021learning} &413&674&5592&27k&1252&1040&7k&888&923&5k \\
	    	\bottomrule
    \end{tabular}	
}
	\caption{Statistics of experimental datasets. Most of our experiments are conducted on these three datasets, where $sp$ represents the seen composition, $up$ indicates the unseen composition, and $i$ is the number of images.}
	\label{table2}
\end{table}
\section{Experiment}
In this section, it will first present the details of experimental datasets and the evaluation metrics. Then, it will provide the technical specifics of the proposed method. The results obtained by our method will be compared with the state-of-the-art methods. Finally, it will analyse the effects of the aforementioned hyperparameters.

\subsection{Datasets and Metrics\label{sec.4.1}}
\textbf{MIT-States} \cite{mit} is a challenging dataset containing 53,753 images. It is annotated to a variety of classes, including everyday objects, with 115 state classes and 245 object classes, and 1,962 compositions in total under the closed-world setting. The number of seen compositions is 1,262 and 700 unseen compositions. 

\textbf{UT-Zappos} \cite{utzappos,yu2017semantic} is a fine-grained dataset consisting of 50,025 images, primarily of various types of shoes, with 12 object classes and 16 state classes, yielding 116 plausible compositions, 83 of which are seen compositions and the rest is unseen. 

\textbf{C-GQA}\cite{naeem2021learning} is the largest dataset for CZSL, containing over 9,000 compositions, including 5,592 seen compositions and 1,932 unseen compositions. It contains 413 state classes and 674 object classes.

\textbf{Evaluation Metrics.} We measure the performance of our method with the following metrics: \textit{1) The accuracy of seen compositions accuracy $\mathcal{A}^{S}$; 2) The accuracy of unseen compositions $\mathcal{A}^{U}$;}; Inspired by the commonly used metrics in Generalized Zero-Shot Learning \cite{xian2017zero}, we calculate the \textit{3) Harmonic Mean $\mathcal{A}^{H}$} to reveal the overall performance of the seen and unseen compositions in an integrated manner, which can be obtained by: $\mathcal{A}^{H}=(2\times \mathcal{A}^{S} \times \mathcal{A}^{U})/ (\mathcal{A}^{U}+\mathcal{A}^{S})$. Considering the case of bias under different operating points, we report the \textit{4) Area Under the Curve (AUC)} to measure the listed method. Finally, we calculate the \textit{5) Accuracy of objects $\mathcal{A}^{obj}$}, and \textit{6) Accuracy of states $\mathcal{A}^{adj}$} referring to \cite{li2022siamese,naeem2021learning} as supplementary metrics to assess the methods.

\textbf{Implementation Details.} For a fair comparison, we employ a fixed ResNet18 \cite{resnet} backbone that is pre-trained on the ImageNet \cite{2009ImageNet} to extract visual features from each image, and the dimension of each visual feature is 512. $E_{s}$, $E_{o}$ and $E_{(s,o)}$ are two fully-connected layers, followed by a ReLU \cite{nair2010rectified} activation function, and the dimension of the output is 512. In our approach, the objects and word vectors are generated from word2vec \cite{mikolov2013distributed}, or fasttext \cite{bojanowski2017enriching} with an additional fully-connected layer in order to project the features into 512 dimensions. Furthermore, the label embedding function can be directly adopted from the existing CZSL method, like Compcos \cite{mancini2021open} (vanilla MLP layers) and CGE \cite{naeem2021learning} (Graph Convolutional Network \cite{kipf2016semi,chen2020simple}). The hyper-parameter of $\gamma$ is set to 1 for MIT-States and UT-Zappos, and 6 for C-GQA.  The adjustable factor $\lambda$ of the objective function is set to 1.5, 1, and 1 for MIT-States, UT-Zappos, and C-GQA, respectively. The learning rate and batch size are set to $5e^{-5}$ and 128 for all three datasets. ADAM optimizer \cite{kingma2014adam} is used to optimize the proposed method. Our model is developed with the PyTorch \cite{paszke2019pytorch} framework and trained on an NVIDIA GTX 2080Ti GPU.
\begin{table*}[t] 
		\centering
		\resizebox{\textwidth}{!}{
			\begin{tabular}{@{}c|cccccc|cccccc|cccccc}
				\toprule
		\multirow{3}{*}{Method} &\multicolumn{6}{c|}{MIT-States} & \multicolumn{6}{c|}{UT-Zappos}& \multicolumn{6}{c}{C-GQA} \\ 
        &\multicolumn{1}{c}{AUC}&\multicolumn{3}{c}{Best}& & &\multicolumn{1}{c}{AUC}&\multicolumn{3}{c}{Best}& & &\multicolumn{1}{c}{AUC}&\multicolumn{3}{c}{Best}& & \\
  &Test&$\mathcal{A}^{H}$&$\mathcal{A}^{S}$&$\mathcal{A}^{U}$&$\mathcal{A}^{adj}$&$\mathcal{A}^{obj}$&Test&$\mathcal{A}^{H}$&$\mathcal{A}^{S}$&$\mathcal{A}^{U}$&$\mathcal{A}^{adj}$&$\mathcal{A}^{obj}$&Test&$\mathcal{A}^{H}$&$\mathcal{A}^{S}$&$\mathcal{A}^{U}$&$\mathcal{A}^{adj}$&$\mathcal{A}^{obj}$\\
				\midrule
        \textit{Base Model}&5.3&18.2&27.9&25.8&28.8&33.2&29.7&44.1&58.4&62.7&44.3&73.6&2.6&12.1&29.7&11.4&9.3&31.9\\
+ ($\mathcal{L}_{o}$,$\mathcal{L}_{s}$)&5.5&18.6&28.2&26.2&28.8&33.3&30.7&45.2&58.8&63.6&44.6&73.8&2.6&12.5&29.0&11.5&10.6&31.3\\
+ $\mathcal{L}_{(s,o)}$&5.9&19.3&30.0&26.3&28.6&33.6&32.3&47.6&59.4&65.2&45.6&73.4&2.8&12.6&29.3&12.3&11.4&30.4\\
\midrule
($\mathcal{L}_{o}$,$\mathcal{L}_{s}$)+$\mathcal{L}_{(s,o)}$&6.0&19.5&30.1&26.3&28.7&33.6 &32.8&48.4&60.7&64.7 &46.3&73.3&2.9&13.1&29.8&12.5&11.5&30.6\\
	\bottomrule
			\end{tabular}
		}
		\caption{Ablation experiments on MIT-States, UT-Zappos and C-GQA.}
	\label{tab.3}
    \end{table*}
    
\subsection{Comparison with the SOTA}
The proposed framework is applied to the vanilla MLP embedding functions (compared to Compcos \cite{mancini2021open}), and the more advanced CGE \cite{naeem2021learning} to show its compatibility with different methods. Specifically, we introduce these embedding functions after the word embedding functions $\varphi_{(s,o)}$ and treat them as additional projection functions. All baseline methods are implemented with their officially released codes.

\textbf{Performence on MIT-State.} MIT-State is a dataset with large inter-class variations and nuisance label noises \cite{atzmon2020causal}. As shown in Tab. \ref{tab.1}, we observe that the performance of the proposed method can significantly outperform the baseline model, including the $AUC$ that are enhanced by $1.1 \% $ and $ 0.9\%$, while $\mathcal{A}^{H}$ of both methods are improved by $2.3\%$. Our method also produces the highest level of results compared to other existing methods, as well as better outcomes than SCEN \cite{li2022siamese}, which employs contrast learning and generative methods. 

\textbf{Performence on UT-Zappos.} The results on UT-Zappos also demonstrate the superiority of the proposed method. We note that the performance gain of $\mathcal{A}^{obj}$ and $\mathcal{A}^{adj}$ by our module provides less substantial than SCEN \cite{li2022siamese} on UT-Zappos. Since the UT-Zappos is a fine-grained dataset, contrastive learning-based approaches such as SCEN are more favorable than our method. However, our approach still can perform better regarding those more critical metrics $\mathcal{A}^{H}$ and $AUC$. This demonstrates that in the situation of insufficient capacity to distinguish a single component, our model better balances the two components, resulting in greater overall accuracy. 

\textbf{Performence on C-GQA.} C-GQA is the relatively most challenging dataset due to its large number of compositions. We achieved the test $AUC$ of $2.9\%, 3.0\%$ outperforming other methods and $\mathcal{A}^{H}$ of $13.1\%$ and $13.6\%$. All metrics have been significantly improved from the baseline models. Nonetheless, the gap between seen and unseen classes in this dataset is still substantial, which might be caused by the complicated compositionality of the C-GQA. 

\subsection{Ablation Study}
\begin{table}[t]
\scalebox{0.76}{
	\centering
    \begin{tabular}{@{}lcccccccc}
    \toprule
		\multirow{2}{*}{Dataset} & \multicolumn{4}{c}{MIT-States}  & \multicolumn{4}{c}{UT-Zappos}  \\
      & $AUC$ & $\mathcal{A}^{H}$   &  \multicolumn{1}{c}{$\mathcal{A}^{S}$}  & $\mathcal{A}^{U}$& $AUC$ & $\mathcal{A}^{H}$   &  \multicolumn{1}{c}{$\mathcal{A}^{S}$}  & $\mathcal{A}^{U}$\\
				\midrule
		\textit{Base Model} &5.3&18.2&27.9&25.8&29.7&44.1&58.4&62.7\\
	  \textit{  + $\mathcal{L}_{F}$}&5.3&18.2&27.6&26.1&31.1&45.6&59.4&64.5 \\
	 \textit{Ours}&6.0&19.5&30.1&26.3&32.8&48.4&60.7&64.7 \\
	    	\bottomrule
    \end{tabular}	
}
	\caption{Comparison of our approach with directly incorporating Focal Loss into the \textit{Base Model}.}
	\label{table4}
\end{table}
Ablation studies are presented to validate the efficacy of our method. Firstly, for the baseline model, we employ a structure that introduces additional classifiers to categorize objects and states on top of CGE \cite{naeem2021learning}, this model is trained with a Cross-Entropy objective function, and the scores of their classifications are then incorporated to the model training to produce the final predictions and this model is denoted as \textit{Base Model}. Then, we added $\mathcal{L}_{o},\mathcal{L}_{s},\mathcal{L}_{(s,o)}$, and denoted them as \textit{Base Model + ($\mathcal{L}_{o},\mathcal{L}_{s})$} and \textit{Base Model +$\mathcal{L}_{(s, o)}$}, respectively. Finally, we denote the complete of our model by \textit{Base Model + ($\mathcal{L}_{o},\mathcal{L}_{s})$+ $\mathcal{L}_{(s, o)}$}. As shown in Tab. \ref{tab.3}, each part added to the \textit{base model} improves the performance on all three datasets. This demonstrates that the two objective functions do not contradict one another but rather complement each other better. 

\textbf{Compared to Focal Loss.} Our method has a formal resemblance to the Focal Loss \cite{lin2017focal} and seeks to address the imbalance issues. However, they are fundamentally different. The primary objective of the Focal Loss is to address the imbalance problem at the class level. In the CZSL scenario, we delineate the performance limitation to the imbalance between the sample's components. To show the superiority of our method, we substituted the Focal Loss with the cross-entropy loss function in the \textit{Base Model} and made a comparison, denoted by \textit{Base Model+$\mathcal{L}_{F}$} in Tab. \ref{table4}. We can observe that simply adding the Focal Loss to the \textit{Base Model} cannot bring immediate improvements to performance on the MIT-States and UT-Zappos.

\subsection{Qualitative Results}

\begin{figure}[ht]
	\centering
	\includegraphics[width=1.00\columnwidth]{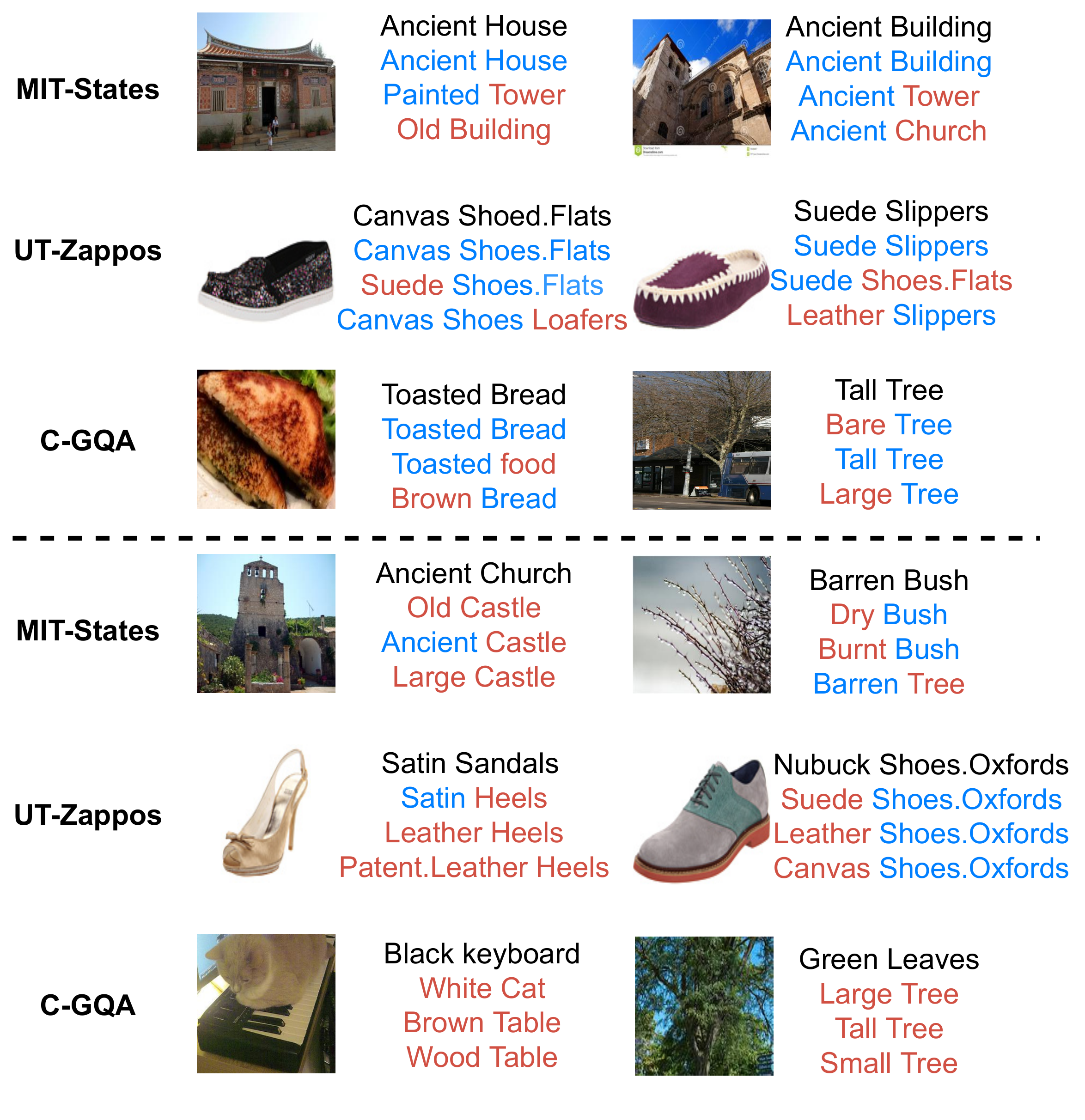} 
	\caption{Qualitative results. We show the top-3 predictions of the model on each of the three datasets, with the first three rows representing correct predictions and the last three rows representing incorrect predictions, correct predictions are shown in blue and errors in red. }
	\label{fig4}
\end{figure}

In addition to the top-1 accuracy, top-3 accuracy can also provide insight into the model's overall performance. Consequently, we present qualitative results for the novel compositions with their top-3 predictions in Fig. \ref{fig4}. The first three rows show examples with accurate predictions, and the last three rows indicate the mispredicted samples, with blue indicating correct and red indicating errors. From the first three rows, we can notice that the dataset requires the model to be optimized toward a unique label, even though most of these samples have acceptable alternative labels.
For example, our model delivers three predictions for ``toasted bread" in the first column of the third row: ``toasted bread," ``toasted food," and ``toasted food." All three labels are appropriate yet not mutually incompatible. This phenomenon demonstrates that our model learns generalizable visual-semantic relationships during training instead of merely seeking to optimize the outcomes. 

The incorrect predictions in the last three rows demonstrate another point. For instance, the sample ``Black Keyboard" in the first column of the sixth row contains three objects: ``cat," ``keyboard," and ``table." From our perspective, it is appropriate for the model to make an error in this situation, given that each sample has only one label provided by subjective manual annotation. In this case, it is challenging for the model to decide whether a ``cat" or ``keyboard" is needed. 

However, because UT-Zappos has finer-grained characteristics than the other two datasets, it is challenging to discriminate between two similar categories using the current structure. For instance, in the second column of the fifth row, the model correctly predicts the object's class, yet the three predictions on the state need to be corrected.

\begin{figure}[t]
	\centering
	\includegraphics[width=1.00\columnwidth]{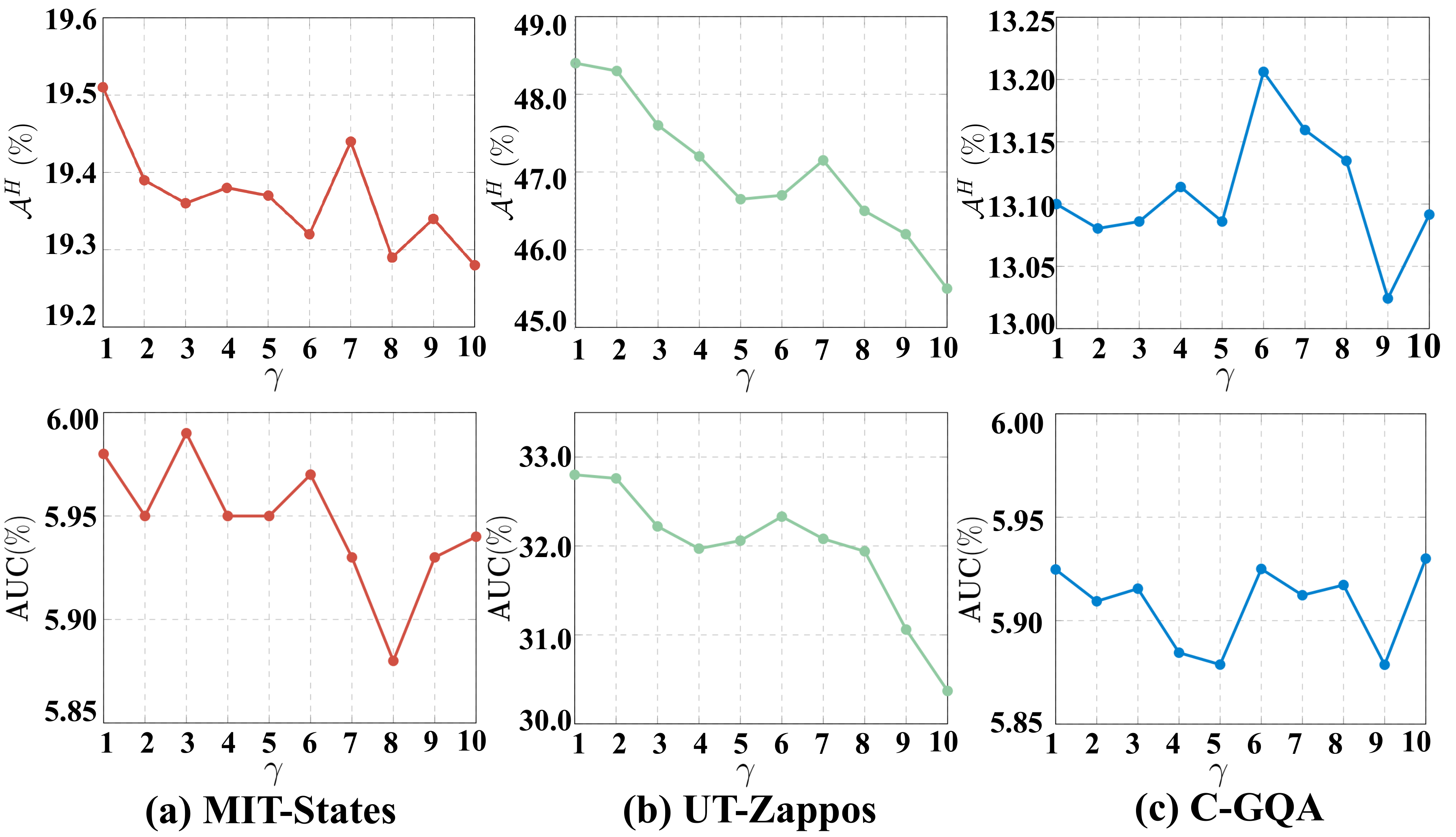} 
	\caption{The influence of the hyper-parameter $\gamma$. }
	\label{fig5}
\end{figure}

\begin{figure}[t]
	\centering
	\includegraphics[width=1.00\columnwidth]{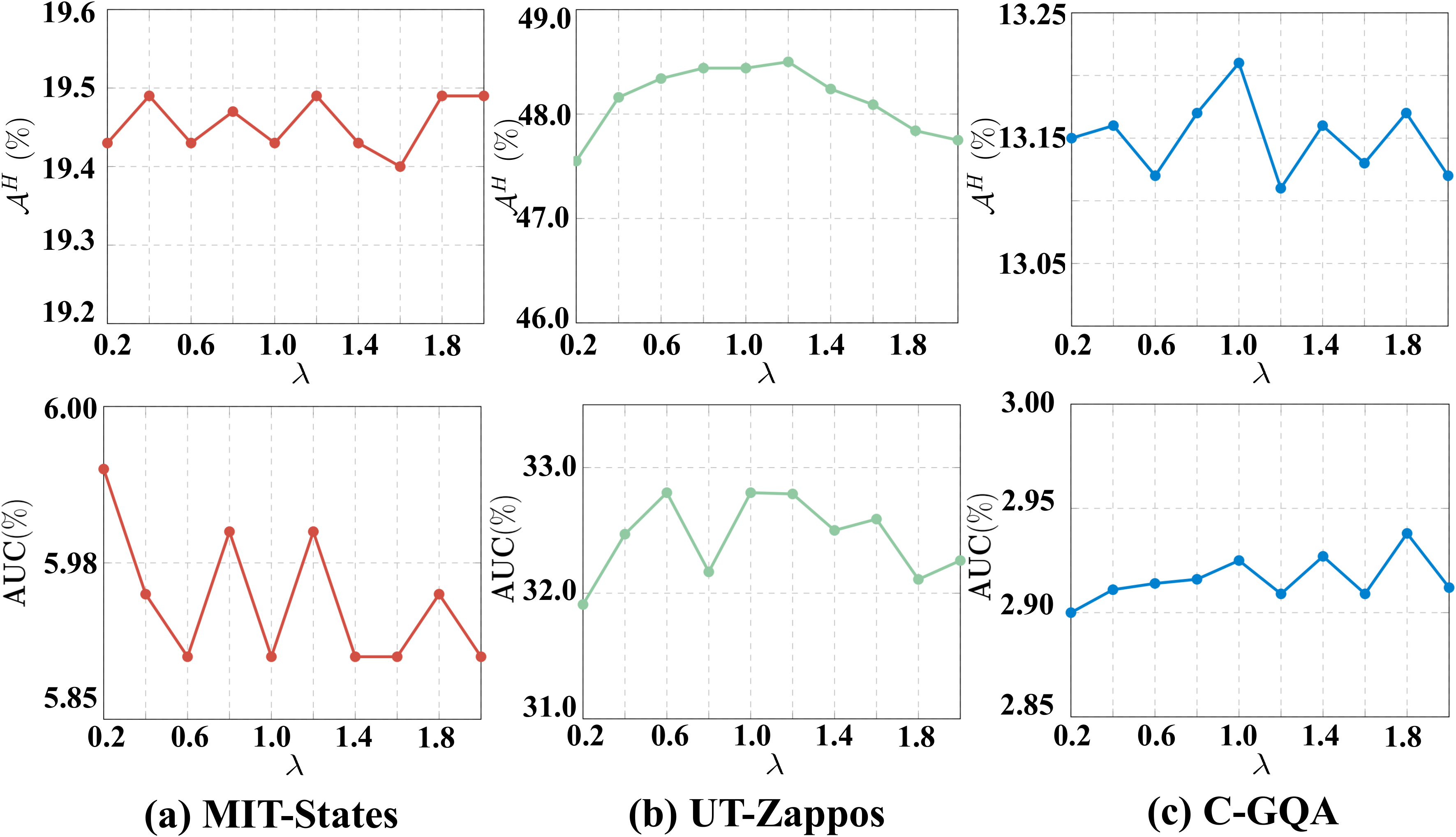} 
	\caption{The influence of the weighting coefficient $\lambda$. }
	\label{fig7}
\end{figure}

\subsection{Hyper-Parameter Analysis}
We focus on analyzing the two hyper-parameters listed below: \textit{1) $\gamma$, the coefficient in Eq. \ref{eq_4}, Eq. \ref{eq_5} and Eq. \ref{eq_7} for adjusting the influence of confidence in objective function}, and \textit{2) $\lambda$ in Eq. \ref{eq_8} for balancing the gradients among the three objective functions.} As shown in Fig. \ref{fig5}, we set $\gamma=1$ , $\gamma=1$, and $\gamma=6$ on MIT-States, UT-Zappos, and C-GQA datasets, and achieved the best results. In Fig. \ref{fig7}, we can observe that the model achieves better performance in synthesis when the $\lambda$ is maintained around 1. Based on the experiments, we presume that the effect of varying parameter values on our model is not significant which in turn illustrates the robustness of our model. Our model's performance on $\mathcal{A}^{H}$ fluctuates between $\pm 0.2\%$, $\pm 3.5\%$ and $\pm 0.2\%$ under the influence of different $\gamma$ settings on the three datasets. And the influence of $\lambda$ on $\mathcal{A}^{H}$ is around $\pm 0.1\%$, $\pm 1.5\%$, and $\pm 0.1\%$ on the three datasets.

\section{Conclusion}
In this research, we examined Compositional Zero-Shot Learning from the standpoint of a multi-label task and characterize its existing task challenges as an imbalance between the two components of the composition. We propose a novel Mutual Balancing in State-Object Components (MUST) method to solve these problems. As a measure of their imbalance, additional modules were developed to examine the visual deviation of the two components in the sample. We utilized this knowledge to balance the components effectively. Our method is compatible with the existing joint embedding function in the current CZSL methods. Adequate experiments demonstrate the effectiveness of our approach, \ie, achieved state-of-the-art on three challenging datasets.

%%%%%%%%% REFERENCEa
{\small
\bibliographystyle{ieee_fullname}

\bibliography{egbib}
}

\appendix
\renewcommand\thefigure{A.\arabic{figure}} 
\renewcommand\theequation{A.\arabic{equation}} 
\renewcommand\thetable{A.\arabic{table}} 
\renewcommand\thesection{\Alph{section}}
\renewcommand\thesubsection{\alph{subsection}}
\clearpage
\section*{Appendix: Mutual Balancing in State-Object Components for Compositional Zero-Shot Learning } 
In this document, we elaborate a bit more on:
\begin{itemize}
\item More about Eq. \textcolor{red}{7}, Eq. \textcolor{red}{8}.
\item Analysis of $\mathcal{L}_{o}$ and $\mathcal{L}_{s}$.
\item Additional experimental results.
\end{itemize}

Our codes are available on \url{https://anonymous.4open.science/r/MUST_CGE/}
\section{More Details about Inference}
We can observe from Tab. \textcolor{red}{3} that $\mathcal{L}_{o}$, $\mathcal{L}_{s}$ and $\mathcal{L}_{(s,o)}$ do not directly enhance the results of $\mathcal{A}^{adj}$ and $\mathcal{A}^{obj}$, Instead, $Base Model$ has improved on the results compared to CGE \cite{naeem2021learning}. Its major difference is the introduction of Eq. \textcolor{red}{7} and Eq. \textcolor{red}{8} in the classification results. Due to the occurrence of visual deviations in the components, we believe it is impractical to expect the two cosine similarities of $d_{s}$ and $d_{o}$ with the components to reach their objectives simultaneously and a similar conclusion can be derived from Fig. \textcolor{red}{2}. Nonetheless, we seek to make further use of this information. Therefore, we introduce Eq. \textcolor{red}{7} and Eq. \textcolor{red}{8}.

\subsection{Ablation Experiments on Inference Methods \label{sec.1}}

\begin{table}[ht]
\scalebox{0.72}{
	\centering
    \begin{tabular}{@{}l|ccc|ccc|ccc}
    \toprule
		\multirow{2}{*}{Methods} & \multicolumn{3}{c|}{MIT-States \cite{mit}}  & \multicolumn{3}{c|}{UT-Zappos \cite{utzappos}} &\multicolumn{3}{c}{C-GQA \cite{naeem2021learning}} \\
      &$\mathcal{A}^{H}$&$\mathcal{A}^{adj}$&$\mathcal{A}^{obj}$ &$\mathcal{A}^{H}$&$\mathcal{A}^{adj}$&$\mathcal{A}^{obj}$ &$\mathcal{A}^{H}$&$\mathcal{A}^{adj}$&$\mathcal{A}^{obj}$  \\
				\midrule
	$Base$&19.3&28.2&32.3&47.2&45.1&73.2&12.8&10.8&29.6\\
	 $Max$& 19.4&28.6&\textbf{33.9}&47.2&45.8&\textbf{75.1}&12.9&11.3&\textbf{31.7}\\
    $Equal$& 19.4&28.6&33.7&47.6&45.6&73.7&13.0&\textbf{11.5}&30.1\\
    \textbf{MUST}&\textbf{19.5}&\textbf{28.7}&33.6&\textbf{48.4}&\textbf{46.3}&73.3&\textbf{13.1}&\textbf{11.5}&30.6\\
	    	\bottomrule
    \end{tabular}	
}
	\caption{$\mathcal{A}^{H}$, $\mathcal{A}^{adj}$, and $\mathcal{A}^{obj}$ under different inference methods, the best results are shown in bold.}
	\label{table1}
\end{table}

With Eq. \textcolor{red}{7}, we adjusted the weighting factor of $d_{s}$ and $d_{o}$ depending on their confidence levels to achieve our objective of excluding components with a lower degree of visual deviations, hence allowing the model to circumvent compositionality's impact in the classification. To prove this conclusion, we designed the following experiment: (1) Directly use $d_{(s,o)}$ as the prediction result, denoted by $Base$, the inference process is: 

\begin{equation}\label{eq_2}
 \resizebox{0.4\linewidth}{!}{$
    \begin{aligned}
  \tilde{y}=\underset{(s,o)\in \mathcal{C}_{closed}}{argmax}d_{(s,o)}.
    \end{aligned}
   $}
\end{equation}

(2) The maximum value in $d_{s}$ and $d_{o}$ is added to the predicted result, denoted by $Max$, the inference process is: 

\begin{equation}\label{eq_3a}
 \resizebox{0.75\linewidth}{!}{$
    \begin{aligned}
  \tilde{y}=\underset{(s,o)\in \mathcal{C}_{closed}}{argmax}[max(\hat{d}_{s}^{(s,o)}, \hat{d}_{o}^{(s,o)})+d_{(s,o)}].
    \end{aligned}
   $}
\end{equation}

(3) $d_{s}$ and $d_{o}$ are added with equal weight to $d_{(s,o)}$, denoted by $Equal$, the inference process is: 

\begin{equation}\label{eq_4a}
 \resizebox{0.75\linewidth}{!}{$
    \begin{aligned}
  \tilde{y}=\underset{(s,o)\in \mathcal{C}_{closed}}{argmax}[\frac{1}{2}(\hat{d}_{s}^{(s,o)}+\hat{d}_{o}^{(s,o)})+d_{(s,o)}].
    \end{aligned}
   $}
\end{equation}

 (4) Finally, we use the results of the method described in the text for ease of observation, denoted by \textbf{MUST}. 
 
 As demonstrated in Tab \ref{table1}, the $Base$ approach performs less efficiently than the other three methods. The primary distinction between the remaining three methods is between inference methods. We can note that the $Max$ method is more advantageous for $\mathcal{A}^{adj}$ and $\mathcal{A}^{obj}$ instances that are more likely to be accurately classified. In contrast, \textbf{MUST} focuses more on achieving a balance between the correctness of both metrics; it also gives a better incentive for the more critical measure $\mathcal{A}^{H}$.
We were able to conclude that $d_s, d_o$'s additional information has some effect on classification, and depending on the weighting method, it can improve the overall results in various aspects. 
\subsection{Further Analysis on Inference Methods}
\begin{table}[ht]
	\centering
\scalebox{0.80}{
	\centering
    \begin{tabular}{@{}l|ccc|ccc}
    \toprule
		\multirow{2}{*}{Datasets} & \multicolumn{3}{c|}{Best Hyper-parameters}  & \multicolumn{3}{c}{Results}\\
      &$\alpha$&$\beta$&$\alpha+\beta$ &$\mathcal{A}^{H}$&$\mathcal{A}^{adj}$&$\mathcal{A}^{obj}$ \\
				\midrule
	MIT-States&0.4&0.6&1.0&19.4&28.6&33.8\\
	UT-Zappos&0.8&0.2&1.0&48.3&46.0&73.1\\
    C-GQA& 0.8&0.1&0.9&13.1&13.2&28.0\\
	    	\bottomrule
    \end{tabular}	
}
	\caption{$\mathcal{A}^{H}$, $\mathcal{A}^{adj}$, and $\mathcal{A}^{obj}$ of using manually defined hyper-parameters for inference instead of Eq. \textcolor{red}{7}.}
	\label{table2a}
\end{table}
To further analyze the role played by Eq. \textcolor{red}{7} in the inference process, we conducted additional experiments. We build another model that is closest to the current \textbf{MUST} approach, \ie, rewriting the Eq. \textcolor{red}{8} as:

\begin{equation}\label{eq_1}
 \resizebox{0.78\linewidth}{!}{$
    \begin{aligned}
  \tilde{y}=\underset{(s,o)\in \mathcal{C}_{closed}}{argmax}[\alpha\hat{d}_{s}^{(s,o)}+\beta\hat{d}_{o}^{(s,o)}+d_{(s,o)}],
    \end{aligned}
   $}
\end{equation}
we directly replace $\omega $ in Eq. \textcolor{red}{8}  with $\alpha$ and $\beta$, and then find the value of $\alpha$ and $\beta$ that achieves the highest result by cross-validation. This form between $Max$ and $Equal$ in Sec. \ref{sec.1} closely resembles our existing strategy. However, there are still distinctions in the finer details, such as it utilize the same weights and are not sample-specific. At the same time, this approach also introduces two additional hyper-parameters, which further increases the cost of model training. We present the best hyper-parameters obtained with their test results in the Tab. \ref{table2a}, we can find two phenomena in it that match our Eq. \textcolor{red}{7}: 1) the sum of $\alpha$ and $\beta$ is best distributed around $1$, and 2) when attempting to get the ideal outcome, $\alpha$ and $\beta$ values are typically not equal. We believe that our method's superior performance is better fitting the two hyper-parameters and contributes to higher flexibility in altering the sample-level weights. 

\section{Further Discussion on the Objective Functions}

\begin{table}[ht]
\centering
\scalebox{0.85}{
	\centering
    \begin{tabular}{@{}l|cc|cc|cc}
    \toprule
	\multirow{2}{*}{Methods} & \multicolumn{2}{c|}{MIT-States}  & \multicolumn{2}{c|}{UT-Zappos}& \multicolumn{2}{c}{C-GQA}\\
   &AUC&  $\mathcal{A}^{H}$ &AUC&  $\mathcal{A}^{H}$ &AUC&  $\mathcal{A}^{H}$\\
				\midrule
	CGE \cite{naeem2021learning}&5.1&17.2&26.4&41.2&2.3&11.4\\
	 +$\mathcal{L}_{(s,o)}$&5.4&18.6&31.2&46.1&2.6&12.6\\
     +$\mathcal{L}_{o/s}+\mathcal{L}_{(s,o)}$&5.8&19.3&32.7&47.2&2.8&12.8\\
	    	\bottomrule
    \end{tabular}	
}
	\caption{Further ablation experiments on the objective functions.}
	\label{table7}
\end{table}

$\mathcal{L}_{s}$ and $\mathcal{L}_{o}$ mainly affect two subsequent parts, (1) the objective function $\mathcal{L}_{(s,o)}$, and (2) the inference process. In the ablation experiments of Sec. \textcolor{red}{4.3}, we have proved that these two objective functions combined with Eq. \textcolor{red}{8} and Eq. \textcolor{red}{7} could maintain a positive model optimization, \ie, $\mathcal{L}_{s}$ and $\mathcal{L}_{o}$ can be considered to have a motivating effect on the inference process of MUST. However, the role of $\mathcal{L}_{s}$ and $\mathcal{L}_{o}$ has not been verified if only the common inference approach is taken, \ie, the classification is done directly using $d_{(s,o)}$. We wish to assess whether $\mathcal{L}_{s}, \mathcal{L}_{o}$ and $\mathcal{L}_{(s,o)}$ have a mutually beneficial relationship to confirm the influence of $\mathcal{L}_{s}$ and $\mathcal{L}_{o}$ on $\mathcal{L}_{(s,o)}$ in this situation, we have designed ablation experiments. We replace the inference method of the $Base Model$ in Sec. \textcolor{red}{4.3} with Eq. \ref{eq_2}, this way the base model is consistent with the CGE \cite{naeem2021learning}, on this basis, we introduce $\mathcal{L}_{(s,o)}$, denoted by CGE+$\mathcal{L}_{(s,o)}$, and at the same time we introduce $\mathcal{L}_o, \mathcal{L}_s$ on this basis, denoted by CGE+$\mathcal{L}_{o/s}$+$\mathcal{L}_{(s,o)}$, to verify the effect played by $\mathcal{L}_o, \mathcal{L}_s$ on $\mathcal{L}_{(s,o)}$.

As shown in Tab. \ref{table7}, we present the test results of the abovementioned approach on various datasets.
We observe that the combination of $\mathcal{L}_o, \mathcal{L}_s$ and $\mathcal{L}_{(s,o)}$ can also provide positive outcomes, and this verifies the role played by $\mathcal{L}_o, \mathcal{L}_s$ as we mentioned in Sec. \textcolor{red}{3.3}.
\section{More Results with Different Settings}
In this section, we will demonstrate the performance of our model with different settings. 

\subsection{Results with Top-1, Top-2, Top-3 Settings}

\begin{table}[ht]
\scalebox{0.66}{
	\centering
    \begin{tabular}{@{}l|ccc|ccc|ccc}
    \toprule
		\multirow{2}{*}{Methods} & \multicolumn{3}{c|}{MIT-States}  & \multicolumn{3}{c|}{UT-Zappos} &\multicolumn{3}{c}{C-GQA} \\
      &1& 2  & 3  & 1& 2&3& 1& 2& 3  \\
				\midrule
     Compcos \cite{mancini2021open} &16.4&25.8&33.0&43.1&61.6&74.3&12.4&18.8&22.8\\
    Compcos + \textbf{MUST}&18.7&27.5&34.2&47.5&66.0&77.1&13.6&20.0&23.9\\
    \midrule
		CGE \cite{naeem2021learning}&17.2&28.5&35.8&41.2&62.8&75.3&11.4&18.3&22.5 \\
	  CGE+ \textbf{MUST}&19.5&30.2&36.8&48.4&66.5&77.3&13.1&19.3&23.9\\
	    	\bottomrule
    \end{tabular}	
}
	\caption{$\mathcal{A}^{H}$ of different datasets under top-1,top-2,top-3 settings, where $1,2,3$ denote top-1, top-2 and top-3 respectively.}
	\label{table3}
\end{table}

\begin{table}[ht]
	\centering
\scalebox{0.80}{
	\centering
    \begin{tabular}{@{}l|ccc|ccc}
    \toprule
		\multirow{2}{*}{Methods} & \multicolumn{3}{c|}{Val AUC}  & \multicolumn{3}{c}{Test AUC}\\
      & 1& 2   & 3  & 1& 2 & 3 \\
				\midrule
	Compcos \cite{mancini2021open} &5.9&13.1&19.4&4.5&10.6&15.9\\
    Compcos + \textbf{MUST}&6.7&14.1&20.2&5.6&11.7&17.2\\
      \midrule
	 		CGE \cite{naeem2021learning}&6.5&15.2&21.6&5.1&12.7&18.1\\
	  CGE+ \textbf{MUST}&7.2&15.4&22.0&6.0&13.3&19.1 \\
	    	\bottomrule
    \end{tabular}	
}
	\caption{Val AUC and Test AUC on MIT-States, where $1,2,3$ denote top-1, top-2 and top-3 respectively.}
	\label{table4a}
\end{table}

Considering the significance of the top predicted results for evaluating CZSL's performance, we also tested this setting. We selected the metrics under top-1, top-2, and top-3 settings for comparison. As shown in Tab. \ref{table3}, our method incentivizes the base model's performance at all scales. Considering that the labels of the samples in CZSL are partially non-mutually exclusive, for example, ``toasted bread" can also be ``toasted food," we feel that the performance under these top-2 and top-3 settings is significant. Tab. \ref{table4a} displays the validation set AUC and the test set AUC for various parameters, and our approach exhibits a similar tendency. 

\subsection{Effect of Different Visual Features on the Results}

\begin{table}[ht]
\scalebox{0.71}{
	\centering
    \begin{tabular}{@{}ll|cc|cc|cc}
    \toprule
	\multirow{2}{*}{Backbone} &\multirow{2}{*}{Methods} & \multicolumn{2}{c|}{MIT-States}  & \multicolumn{2}{c|}{UT-Zappos}& \multicolumn{2}{c}{C-GQA}\\
   &   &AUC&  $\mathcal{A}^{H}$ &AUC&  $\mathcal{A}^{H}$ &AUC&  $\mathcal{A}^{H}$\\
				\midrule
	\multirow{2}{*}{ResNet50 \cite{resnet}} &	CGE \cite{naeem2021learning}&6.7&20.5&29.3&43.7&3.4&14.4\\
	 & CGE+ \textbf{MUST}&7.2&21.3&34.2&49.2&4.1&15.8 \\
     \midrule
     \multirow{2}{*}{ResNet101 \cite{resnet}} &	CGE \cite{naeem2021learning}&6.9&20.8&26.2&40.8&3.4&14.1\\
	 & CGE+ \textbf{MUST}&7.3&21.3&32.7&47.8&3.8&15.3 \\
	    	\bottomrule
    \end{tabular}	
}
	\caption{Results with different backbones.}
	\label{table5}
\end{table}

To determine the effect of various visual features on our approach, we substitute the model's backbone with ResNets \cite{resnet} of varying depths. These backbones are pre-trained in ImageNet \cite{2009ImageNet} and are fixed during our model's training. As shown in Tab. \ref{table5}, we evaluated the model's performance using ResNet50 and ResNet101 as the backbone, where the CGE results are generated from officially released codes. CGE produces a floating result after replacing the backbone, as the empirically deeper backbone can provide different discriminating information about the visual features. But our model can continue to incentivize the base model, which has proven the robustness of our approach.

\end{document}